\documentclass[sigconf, authorversion]{acmart}
\AtBeginDocument{%
  \providecommand\BibTeX{{%
    \normalfont B\kern-0.5em{\scshape i\kern-0.25em b}\kern-0.8em\TeX}}}

\hypersetup{
           breaklinks=true,   
           colorlinks=true,   
        }
\copyrightyear{2022} 
\acmYear{2022} 
\setcopyright{acmlicensed}
\acmConference[SIGSPATIAL '22]{The 30th International Conference on Advances in Geographic Information Systems}{November 1--4, 2022}{Seattle, WA, USA}
\acmBooktitle{The 30th International Conference on Advances in Geographic Information Systems (SIGSPATIAL '22), November 1--4, 2022, Seattle, WA, USA}
\acmPrice{15.00}
\acmDOI{10.1145/3557915.3561473}
\acmISBN{978-1-4503-9529-8/22/11}

\begin{document}

\title{Vision Paper: Causal Inference for Interpretable and Robust Machine Learning in Mobility Analysis}

\author{Yanan Xin}
\orcid{}
\affiliation{%
  \institution{ETH Z{\"u}rich, Institute of Cartography and Geoinformation}
  \streetaddress{Stefano-Franscini-Platz 5}
  \city{Z{\"u}rich}
  \country{Switzerland}
  \postcode{8093}
}
\email{yanxin@ethz.ch}

\author{Natasa Tagasovska}
\affiliation{%
  \institution{Prescient Design, Genentech | Roche}
  \streetaddress{Grenzacherstrasse 124}
  \city{Basel}
  \country{Switzerland}
  \postcode{4070}
}
\email{natasa.tagasovska@roche.com}

\author{Fernando Perez-Cruz}
\affiliation{%
  \institution{Swiss Data Science Center}
  \streetaddress{Turnerstrasse 1}
  \city{Z{\"u}rich}
  \country{Switzerland}
  \postcode{8092}
}
\email{fernando.perezcruz@sdsc.ethz.ch}

\author{Martin Raubal}
\affiliation{%
  \institution{ETH Z{\"u}rich, Institute of Cartography and Geoinformation}
  \streetaddress{Stefano-Franscini-Platz 5}
  \city{Z{\"u}rich}
  \country{Switzerland}
  \postcode{8093}
}
\email{mraubal@ethz.ch}

\renewcommand{\shortauthors}{Xin, et al.}

\begin{abstract}
  Artificial intelligence (AI) is revolutionizing many areas of our lives, leading a new era of technological advancement. Particularly, the transportation sector would benefit from the progress in AI and advance the development of intelligent transportation systems. Building intelligent transportation systems requires an intricate combination of artificial intelligence and mobility analysis. The past few years have seen rapid development in transportation applications using advanced deep neural networks. However, such deep neural networks are difficult to interpret and lack robustness, which slows the deployment of these AI-powered algorithms in practice. To improve their usability, increasing research efforts have been devoted to developing interpretable and robust machine learning methods, among which the causal inference approach recently gained traction as it provides interpretable and actionable information. Moreover, most of these methods are developed for image or sequential data which do not satisfy specific requirements of mobility data analysis. This vision paper emphasizes research challenges in deep learning-based mobility analysis that require interpretability and robustness, summarizes recent developments in using causal inference for improving the interpretability and robustness of machine learning methods, and highlights opportunities in developing causally-enabled machine learning models tailored for mobility analysis. This research direction will make AI in the transportation sector more interpretable and reliable, thus contributing to safer, more efficient, and more sustainable future transportation systems.
  
\end{abstract}



\maketitle

\section{Advancements and Challenges in Deep Learning-Based Mobility Analysis}
Urban mobility has been rapidly increasing in volume as the progress of urbanization accelerates. Despite the advantages and opportunities mobility has brought to our society, there are also severe drawbacks such as the transport sector’s role as one of the main contributors to greenhouse-gas emissions. These drawbacks have posed great challenges to achieving several of the Sustainable Development Goals as formulated by the United Nations Development Programme (UNDP, 2015), including good health and well-being, sustainable cities and communities, and climate action \cite{raubal2020spatial}. 

Mobility data analysis plays an integral part in addressing these issues, supported by the research on developing innovative spatial and computational methods \cite{raubal2020spatial}. Recent research on computational methods for mobility analysis has given increasing attention to black-box deep learning methods, because of their superior predictive power compared to traditional methods in many mobility-related applications. Some of these applications include but are not limited to traffic forecasting, travel demand prediction, traffic signal control, traffic incident inference, transport mode detection, and traffic congestion detection. Detailed discussions of these applications can be found in multiple survey papers \cite{veres2019deep, luca2021survey, kumar2021applications}.

Admittedly, achieving accurate predictions using these deep learning methods is sufficient in certain use cases. However, a lack of interpretability and robustness can substantially slow the deployment of these AI-powered algorithms in practice \cite{lana2021data}. Beyond predicting mobility patterns, it is more useful to interpret these predictions so that we can understand what factors influence them and how we can apply policy interventions to achieve more desirable outcomes, such as avoiding a traffic jam. Besides, models need to be robust against dynamic changes in the deployment environment, for example, a change in mobility patterns over time, and be able to detect such changes for model updates. 

The lack of robustness undermines the reliability of a model and makes the system vulnerable to unforeseen input or adversarial attacks. Furthermore, the decision-making process of many deep learning models is unclear to both model developers and end users, making it difficult to derive actionable information that could be used for policymaking and to keep the public informed of the rationale behind the decisions. The need to improve the interpretability and robustness of machine learning methods is also manifested by the growing number of governmental regulations that demand an explanation for machine-derived decisions \cite{carvalho2019machine}.

Research on interpretable and robust machine learning has gained increasing attention in the last few years, proliferating multiple methodologies \cite{carvalho2019machine, molnar2020interpretable}. Among these methods, the intersection of causal inference and machine learning has recently gained traction and advanced significantly. According to Judea Pearl's three-level causal hierarchy, i.e., association, intervention, and counterfactual, machine learning methods are limited to the association level of the causal hierarchy \cite{pearl2019seven}, since they rely purely on associative relationships extracted from observational data. To address the issues of model explainability and robustness, higher levels of causal reasoning (i.e., intervention and counterfactual) are indispensable. Causal relationships are by definition interpretable and invariant. Causal links offer explanations of the model and/or its predictions; and underlying causal structure holds true across different problem setups and environments, thus can be leveraged to improve the transferability and robustness of machine learning models.

Although many interpretable and robust machine learning methods have been developed, most of these are tailored for image data and sequential data \cite{carvalho2019machine, rojat2021explainable}, which cannot address certain challenges of mobility data analysis. The main challenge of applying known causal modeling approaches to mobility data or in general spatial data is that the spatiotemporal dependency often violates the independent observations assumption used by many causal discovery and inference approaches, such as Rubin’s causal model, fast causal inference algorithm, invariant causal prediction, etc. \cite{akbari2021spatial, mielke2020discovering}. Besides, the large sample size, dimensionality (space, time, attributes), and dynamic properties of mobility data also pose great challenges in inferring causal effects. Some efforts have been devoted to developing spatially-explicit causal inference approaches, such as the spatial difference-in-differences approach \cite{dube2014spatial} and the current fast causal inference method \cite{mielke2020discovering}. Although those causal methods are not designed for machine learning models, they provide good starting points in developing causally-enabled machine learning methods for spatiotemporal data. In addition, a few recent studies used causality-inspired machine learning methods for agent-based motion forecasting, and showed that utilizing causality holds great promises in improving generalization and interpretability \cite{mcduff2022causalcity, liu2022towards}. Overall, leveraging and adapting causal models for mobility data analysis, especially in the combination of complex machine learning models, is still in its very infancy.

\section{Causal Inference for Interpretable and Robust Machine Learning}

\subsection{Causality for Interpretability}
To further improve the applicability of data-driven models across industries and domain sciences, the machine learning community shows an increasing effort in improving model robustness, reliability, and interpretability. This tendency triggered a line of research that discovered pitfalls in terms of inability to recognise an out-of-distribution data example due to the lack of uncertainty quantification \cite{lakshminarayanan_simple_2017, tagasovska_single-model_2019}, as well as inability to handle and generalize to distributional shifts \cite{ganin_domain-adversarial_2016}. Hence, these motivated a significant interest in how humans and machines perceive/understand the world and how methods derive their predictions, namely, aiming for causal reasoning in machine learning. Incorporating causal knowledge into data-driven approaches has been of interest since the 1980s, when Judea Pearl \cite{pearl_probabilistic_1988} pioneered the fusion of graphical models and structural equations, providing for the first time a mathematical framework for causal discovery and inference \cite{pearl_causality_2009}. This introduced a new subfield in machine learning with a focus on learning graphical models from observational and interventional data \cite{peters_elements_2017, scholkopf_causality_2019}. However, such methods have recently hit a barrier due to overly intense computational operations (exponentially increasing subsets of variables to be evaluated as potential causal models), as well as a lack of ground truth reference data in real-life scenarios. This follows from the fact that a trustworthy causal model should be validated by one, if not multiple domain experts \cite{mooij_distinguishing_2016}. Nevertheless, the causal framework has continued its quest of improving machine learning models, taking a different turn, by leveraging different interventional distributions and using the concept of counterfactuals for interpretability. However, since causal inference is very ambitious and controversial (due to subjective views), these techniques are thought as ``geared towards causality'' but not necessarily able to infer the underlying true causal effects \cite{buhlmann_invariance_2020}. Still, they do something more intelligent aiming towards causality than an analysis based on a standard statistical model and often have more valuable interpretation than standard machine learning methods. As such they have an important role in the crucial quest for interpretable machine learning.

\subsection{Causality for Robustness}
A widely shared view of robustness in modelling is perceived through invariance or stability \cite{peters_causal_2016,  buhlmann_invariance_2020}. Causal relationships are by definition invariant, meaning they hold true across different circumstances and domains. This is a desirable property for machine learning systems, where we often predict on data that we have not seen in training; we need these systems to be adaptable and robust. Peters et al. \cite{peters_causal_2016} have proposed to exploit the invariance of causal models for machine learning inference. The key underlying idea is to gather all sub-models that are statistically invariant across environments (e.g., different geographical locations or time periods, etc.) in a suitable sense. The causal sub-model, with a direct causal effect on the target variable, will be one of these invariant sub-models. As such, Peters et al. \cite{peters_causal_2016} introduced a new approach to make use of causal inference in methods that exploit the invariance principle to propose: an optimization framework that under mild assumptions guarantees generalization \cite{arjovsky_invariant_2019}, improved reinforcement learning frameworks \cite{ahuja_invariant_2020}, as well as leveraging it for domain adaptation and transfer learning \cite{rojas-carulla_invariant_2018}.

\section{Opportunities for Causally-Enabled Machine Learning Methods in Mobility Analysis}

In this section, we emphasize key challenges of deep learning-based mobility analysis that require interpretability and robustness, and highlight opportunities for causally-enabled machine learning approaches to address those challenges.

\textbf{Opportunity 1: Identify influences of mobility data representation on predictions.}

A critical challenge of deep learning-based mobility analysis has been the lack of understanding of how data representation influences the predictions. Mobility data can be represented in different formats, spatial, and temporal scales due to the many data collection methods and various needs of transportation applications. The representation of mobility data has been shown to have a key impact on the analytical results, such as the predictability of individual mobility \cite{qian2013impact}, characterization of aggregate mobility behavior \cite{paul2016scaling}, transport mode detection \cite{burkhard2020requirements}, and traffic forecasting \cite{ermagun2018spatiotemporal}.

The impact of such data representation on model predictions has been examined for mobility analysis using more traditional machine learning models \cite{burkhard2020requirements}, but little is known about mobility analysis using deep learning models. Causal intervention can offer insights into the sensitivity of deep learning models towards different data representations through carefully-designed interventions. Based on mobility rules, we can design causal graphical models and simulate distributional changes per single or multiple variables  \cite{peters_causal_2016}. These changes encode the representational change and allow us to systematically evaluate the influence of these changes on the predictions of deep neural networks. These understandings are critical in developing benchmark mobility data sets and evaluation metrics to compare different deep learning methods used for mobility analysis. Moreover, the understanding of data representation also provides crucial data specifications that improve the robustness and interoperability of a model.

\textbf{Opportunity 2: Evaluate causal effects of input features and derive actionable insights.}

The second critical challenge is the lack of methods for identifying the causal effects of input features used in deep learning models. Since mobility data are context-dependent, contextual features have shown to play a key role in mobility applications \cite{andrienko2011event}. However, feature selection can be time-consuming and sometimes requires domain knowledge, and thus more contextual variables than needed are often used as input to machine learning models for mobility modelling. Deep learning models excel in capturing statistical correlations and complex relationships among variables, thus often producing state-of-the-art predictive performance. Yet, such prediction is still based on statistical correlations, which alone cannot identify causes of an event. This offers little actionable information for intervention to avoid an undesirable outcome, such as a traffic incident or traffic congestion \cite{lana2021data}. Besides, due to the complex structure of deep learning models, it is difficult to evaluate the importance of a feature, which makes it challenging to conduct feature selection. 

As a promising solution, causality can be used to guide the selection of explainable and robust features \cite{yu2020causality}.  Besides, counterfactual reasoning can provide explanations for black-box models retrospectively \cite{ wachter_counterfactual_2018} and offer suggestions to actions that lead to desirable alternative outcomes. For a prediction for a certain input, we can generate multiple minimum perturbed samples of that input that cause the prediction value to change. This will give us an insight, for example in the application of traffic forecasting, into how we could have changed the on-ramp flow (e.g., adding ramp metering) such that the volume of traffic at that time point would have been reduced. 

\textbf{Opportunity 3: Increase robustness of deep neural networks against spatio-temporal domain shifts.}

Mobility data constantly change over space and time, thus data used during test time might come from a different underlying distribution than the training data, resulting in a discrepancy between the training and testing data distributions, called a domain shift. When domain shifts happen, the model will likely fail in maintaining its performance, which calls for further updates of the model. It is critical to identify domain shifts in time to minimize the potential loss or damage caused by the poor performance of the model, and to improve the reliability of the model when transferring it across different geographic regions or time periods. 

Leveraging the invariance nature of causal features, we can identify domain-invariant features to improve the robustness to spatio-temporal domain shifts. For example, Invariant Causal Prediction methods can be used to distinguish between so-called core/causal features and style/spurious ones \cite{peters_causal_2016, heinze-deml_invariant_2018}. In the case of traffic forecasting, one could expect road density to be a core feature while the density of Points of Interest only projects spurious dependence on traffic volume due to its strong correlation with road density. Having sufficient and diverse data will allow us to distinguish the causal features by appearing as significant predictors across multiple environments when predicting the same target.

\textbf{Opportunity 4: Increase robustness of deep neural networks against geometric change.}

Raw mobility data are sequences of time-stamped geolocations depicting the trajectories of humans. These trajectories can be represented as individual linear features or road network graphs. To efficiently store, analyse, and visualize mobility data, trajectory simplification is often applied to reduce redundant information. This is particularly useful for routing and navigation applications in mobile devices that have limited storage and battery capacity \cite{khot2014road}.

Most trajectory simplification algorithms remove points or segments of trajectories based on either the importance of geometrical or semantic information \cite{douglas1973algorithms, richter2012semantic}. When trajectory data are aggregated into road network graphs, map compression can be further used to simplify the road network graph by deleting certain nodes and edges \cite{khot2014road}. During these trajectory simplification or map compression processes, the geometry of trajectory data is inevitably distorted. These trajectory simplification methods do not consider the influence of the geometric change on the predictive accuracy of machine learning models. To ensure that machine learning models keep a desired level of predictive accuracy when different trajectory simplification algorithms are used, domain-invariant/causal representations can be used to guide the simplification process such that the information loss of causal features is minimized. By extracting features that are invariant across different levels of simplifications, we can improve the robustness of neural networks against the geometric change of trajectory data in general, thus making neural networks robust to adversarial attacks.

\section{Conclusion}
In this paper, we envision opportunities of utilizing causal inference to enhance the interpretability and robustness of deep learning methods and address specific challenges in mobility analysis. This research direction will enhance our understanding of factors that influence the performance of machine learning models and help design benchmark data sets and more suitable criteria for evaluating deep learning models used for mobility analysis. It also holds great promises in guiding the development of learning algorithms that are robust to domain shifts. The outcomes can assist transportation planners to evaluate the risk and maturity of deploying a machine learning model in practice. In addition, the research will enrich our understanding of the type of explanations that are technically feasible and help policymakers refine AI regulations (e.g., Ethics Guidelines for Trustworthy Artificial Intelligence \cite{doi/10.2759/346720}) related to mobility applications. 

\begin{acks}
We thank the Hasler Foundation for supporting the research project through grant 1-008062.

\end{acks}

\bibliographystyle{ACM-Reference-Format}
\bibliography{author-version}

\end{document}